\def\year{2021}\relax
\documentclass[letterpaper]{article} % DO NOT CHANGE THIS
\usepackage{aaai21}  % DO NOT CHANGE THIS
\usepackage{times}  % DO NOT CHANGE THIS
\usepackage{helvet} % DO NOT CHANGE THIS
\usepackage{courier}  % DO NOT CHANGE THIS
\usepackage[hyphens]{url}  % DO NOT CHANGE THIS
\usepackage{graphicx} % DO NOT CHANGE THIS
\usepackage{natbib}  % DO NOT CHANGE THIS AND DO NOT ADD ANY OPTIONS TO IT
\usepackage{caption} % DO NOT CHANGE THIS AND DO NOT ADD ANY OPTIONS TO IT
\usepackage{amsthm}
\usepackage{amsmath}
\newtheorem{theorem}{Theorem}
\newtheorem{proposition}{Proposition}
\newtheorem{definition}{Definition}
\newtheorem{example}{Example}
\newtheorem{corollary}{Corollary}

\usepackage[switch]{lineno}  %
\usepackage{transparent}
\usepackage{pgfplots}
\usepackage{amssymb}
\usepackage{tikz}
\usepackage{graphs}
\usepackage{tkz-euclide}
\usetikzlibrary{intersections}
\usetikzlibrary{positioning}
\usetikzlibrary{arrows}
\usetikzlibrary{shapes}
\usetikzlibrary{calc}
\usepackage{pmat} %%%%%% To draw matrices with dashed divisions. Info at: http://tug.ctan.org/macros/plain/pmat/pmat.pdf

\usepackage[inline]{enumitem} % To play with items naming
\usepackage{bm}

\newcounter{questionchapter}
\newcounter{question}[questionchapter]
\newcounter{questionprob}[questionchapter]
\newcounter{questiontime}[questionchapter]
\renewcommand\thequestion{\textbf{Q\arabic{question}}}
\renewcommand\thequestionprob{\textbf{P\arabic{questionprob}}}
\renewcommand\thequestiontime{\textbf{D\arabic{questiontime}}}
\newenvironment{questions}[1][]{\refstepcounter{questionchapter}\refstepcounter{question}\par\medskip
\begin{list}{\thequestion}{\usecounter{question}}}{\end{list}\medskip}

\renewcommand{\inf}{\mu}
\newcommand{\pnomem}{P}
\newcommand{\pmem}{P}

\newcommand{\well}{well-behaved}
\newcommand{\Well}{Well-behaved}

\newcommand{\State}{S}
\newcommand{\memState}{R}
\newcommand{\MState}{\bar{\memState}}
\newcommand{\mstate}{\bar{s}}
\newcommand{\msdis}{\bar{\mu}}
\newcommand{\stochastic}{\{\State_{\round}\}_{\round \geq 0}}
\newcommand{\memstochastic}{\{\memState_{\round}\}_{\round \geq m + 1}}
\newcommand{\mstochastic}{\{\MState_{\round}\}_{\round \geq m}}
\newcommand{\state}{s}
\newcommand{\memstate}{s}

\newcommand{\round}{t}

\newcommand{\Colour}{X}

\newcommand{\maxmemory}{m}
\newcommand{\mG}{\bar{G}}
\newcommand{\mV}{\bar{V}}
\newcommand{\mE}{\bar{E}}
\newcommand{\average}{\tau}
\newcommand{\weight}{w}

\newcommand{\blue}{\text{blue}}
\newcommand{\red}{\text{red}}
\newcommand{\green}{\text{green}}
\newcommand{\torus}{torus}

\newcommand{\textdef}[1]{\textbf{#1}}

\newcommand{\citenp}[1]{\citeauthor{#1} \citeyear{#1}}
\newcommand{\citenptext}[1]{\citeauthor{#1} (\citeyear{#1})}
\newcounter{questionscounter}

\title{The Influence of Memory in Multi-Agent Consensus}
\author {

        David Kohan Marzagão,\textsuperscript{\rm 1}

        Luciana Basualdo Bonatto,\textsuperscript{\rm 2}
        Tiago Madeira,\textsuperscript{\rm 3} \\
        Marcelo Matheus Gauy,\textsuperscript{\rm 3}
        Peter McBurney\textsuperscript{\rm 1} \\
        
}
\affiliations {
    \textsuperscript{\rm 1} King's College London \\
    \textsuperscript{\rm 2} University of Oxford \\
    \textsuperscript{\rm 3} University of São Paulo 
    \\ 
    \{david.kohan, peter.mcburney\}@kcl.ac.uk,  basualdobona@maths.ox.ac.uk, \{madeira, mmgauy\}@ime.usp.br \\

}
\begin{document}

\maketitle

\begin{abstract}
    Multi-agent consensus problems can often be seen as a sequence of autonomous and independent local choices between a finite set of decision options, with each local choice undertaken simultaneously, and with a shared  goal of  achieving  a global consensus state. Being able to estimate probabilities for the different outcomes and to predict how long it takes for a consensus to be formed, if ever, are core issues for such protocols.
    
    Little attention has been given to protocols in which agents can remember past or outdated states.  In this paper, we propose a framework to study what we call \emph{memory consensus protocol}. We show that the employment of memory allows such processes to always converge, as well as, in some scenarios, such as cycles, converge faster. We provide a theoretical analysis of the probability of each option eventually winning such processes based on the initial opinions expressed by agents. Further, we perform experiments to investigate network topologies in which agents benefit from memory on the expected time needed for consensus.
\end{abstract}

\section{Introduction}

Many applications of distributed computing involve autonomous entities making individual, independent assessments of some situation, based only on limited or local knowledge, in repeated decision rounds until a global consensus decision emerges, if it ever does.  The most famous of these applications nowadays is probably the decision-making process used in blockchain or distributed ledger applications, but computational applications long predated the development of Bitcoin in 2008 \cite{nakamoto2008bitcoin,tsitsiklis1984problems,olfati2007consensus}.  Applications continue to emerge, for example, in the design of collective decision-making processes for groups of autonomous robots or drones~\cite{yan2013survey, ismail2018survey}.

Many of these decision processes can be modelled as a process between autonomous agents played on a graph, where the nodes of the graph represent the autonomous entities, and the edges between nodes represent connections or information transfers between these entities.   The outcomes of the decisions are represented by a set of finite states or labels, often called \emph{opinions} or \emph{colours}, which are the possible decision options for each agent at each round of the process. The local nature of agent knowledge is manifested by the topology of the graph, in that nodes are typically only connected to \emph{some} other nodes, and not to all others. Thus, an agent or node may know at the start of each round the states at the previous round of the nodes to which it is connected, and then use this local knowledge to decide what state it should adopt at the current  round. Agent decisions are made synchronously. Synchronous consensus processes have been extensively studied (e.g. \citenptext{martinez2005synchronous}; \citenptext{lynch1996distributed}; \citenptext{cao2015event}; \citenptext{olfati2007consensus}). The protocol will typically assume that all agents have the same desired final goal, 
which is that all nodes choose a particular state, i.e., reach a consensus. These protocols also typically assume that the agents are fungible; in other words, that they all use the same algorithm to decide what state to adopt at each round and are not otherwise internally distinguishable (although they may have different numbers of connections). For the context of this paper, we assume all agents act sincerely, without malice or whimsy.

Core issues for such protocols involve being able to compute the consensus probabilities for each of the different outcomes and to predict how long it takes for a consensus to be formed. A known feature of synchronous consensus processes, however, is that, for some network topologies (e.g., even cycles), they may encounter deadlocks, i.e., configurations from which no consensus can be reached \cite[Sec 2.1]{hassin2001distributed}. Moreover, some network topologies (e.g., odd cycles) may have structure locally similar to the ones where deadlocks are encountered and the consensus process may take a long time to converge as a result. We aim to address these issues by considering that agents may remember and take into account previous rounds in their decision-making. 

This paper asks what are the effects, if any, on the likelihood and speed of convergence of agents having a memory of some past states of the current process.  In what we call \textdef{memory consensus protocols}, agents may copy either their neighbours' past states or their current ones, according to different probabilities. We will show that, with memory, agents are not only able to avoid deadlocks in networks such as even cycles, but also converge in fewer rounds for several graph structures.  We use a mix of probabilistic analysis and simulation to explore these questions.  The main contributions of the paper are: 

    \begin{enumerate}
        \item A framework to analyse the synchronous multi-agent consensus protocol when agents remember previous rounds.  
        \item A theoretical and complete analysis of the probabilities of each colour winning a consensus process with memory given the initial states. We also show that such processses always converge to a consensus.
        \item A comprehensive exploration of different graph structures showing in which situations the employment of memory reduces the expected number of rounds for convergence. 
    \end{enumerate}

\section{Background and Main Definitions}\label{sec:background}
    In this section, we present concepts and results from the literature that will be used in the subsequent sections. We first introduce the classical version of consensus protocol used in this paper, also known as voter model~\cite{donnelly1983finite, nakata1999probabilistic, hassin2001distributed, cooper2016linear}, in which agents have no memory of past rounds. We then propose a definition of stochastic consensus processes in which memory is taken into account. 
    
    \subsection{Memoryless Consensus Protocol}
    The \textdef{memoryless consensus protocol} defines a round-based consensus process on a strongly connected directed graph $G = (V,E)$.\footnote{Henceforth, we assume all graphs are strongly connected unless stated otherwise.} In such processes agents are represented by nodes in this graph. At each round, each node has a colour associated to it, representing the respective agent's current state (or opinion). Their goal is to reach consensus, i.e., a situation where every agent is in the same state. To that end, at each round, all agents update their state synchronously based on the colour of their out-neighbours.\footnote{For precision, we consider that agents change their state at the end of each round, after all nodes have made their decisions.}
    The probability that $v$ copies colour of node $u$ in a given round is represented by the weight of edge $(v,u)$. The weights of edges starting at a given node are assumed to be positive and to sum to $1$. We adopt the notation $\weight(v,u) = 0$ if $(v,u) \notin E$, and note that self loops are allowed and thus $v$ may adopt its own colour. Once reached, a consensus is stable. The term `memoryless' comes from the fact that, at time $t$, agents decide on a colour for time $t+1$ based only on other states at time $t$, and keep no record of previous states (not even their own previous colours). 

    Let $\Colour = \{c_1, \dots, c_k\}$ be the set of all possible colours on a consensus process. A \textdef{configuration} on a graph $G=(V,E)$ is a vector $\state \in \Colour^V$ such that $\state(v)$ represents $v$'s colour in configuration $\state$. Formally, a process is a sequence of random variables $\stochastic$, with  $\State_{t+1}\in \Colour^V$ being a configuration generated based on $\State_t$ and the algorithm described above. We say colour $i$ \textdef{wins} the process if a configuration $\State_t =\state$, such that $\state(v) = i$ for all $v$, is reached.

    \begin{example}\label{exm:memoryless}
        Consider the graph shown in Figure \ref{fig:example_1}, in which $V = \{v_1, v_2, v_3\}$ and $\Colour =\{\red, \blue\}$. Assume it shows a process in its initial state. Thus, $\State_0(v_1) = \State_0(v_2) =  \blue$, whereas $\State_0(v_3) = \red$. 
        The update algorithms are such that $v_2$ will copy $v_1$'s colour w.p. $\frac{1}{4}$, and $v_3$'s w.p. $\frac{3}{4}$. Agent $v_3$, on the other hand, has a probability of $\frac{2}{3}$ of keeping its own colour, otherwise copies $v_2$'s. Finally, $v_1$ behaves deterministically in this graph by always copying $v_2$'s state. 
        
        \begin{figure}
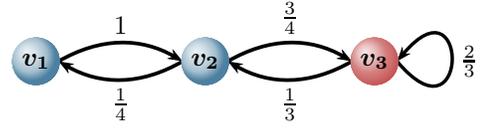

            \centering
            \memlessgraph         
            \caption{A Possible Initial Configuration of a Memoryless Consensus Process on a Graph  $G$.}
            \label{fig:example_1}
        \end{figure}
    \end{example}
    
    There are graphs for which the probability of reaching consensus is not $1$. Such graphs have what can be seen as `deadlocks'. As an extreme example, deadlocks may occur in directed cycles, where there is an edge from every node only to it's clockwise neighbour. Any non-consensus state will generate another non-consensus state in the subsequent round by simply rotating colours anti-clockwise.
    The idea of deadlocks is formalised in Definition~\ref{def:well-behaved}.

    \begin{definition}[{\Well} Graphs]\label{def:well-behaved}
        A graph is said to be {\well} if consensus processes on it reach a consensus with probability $1$ for any initial configuration.
    \end{definition}
    
    Proposition~\ref{prop:gcd}, taken from ~\citenptext{marzagao2017team}, gives the necessary and sufficient conditions for which directed graphs are {\well}. It also applies to undirected graphs by replacing each edge by a pair of antiparallel ones. 

    \begin{proposition}[\citenp{marzagao2017team}]\label{prop:gcd}
        A directed graph $G$ is {\well} if and only if the greatest common divisor (gcd) of the lengths of all cycles in $G$ is equal to 1. 
    \end{proposition}
    \begin{example}\label{exm:badly-behaved}
        Classical examples of undirected graphs that are \emph{not} {\well} are cycles of even length, paths, and trees. More generally, from Proposition \ref{prop:gcd}, an undirected graph is well-behaved if and only if it is not bipartite. 
    \end{example}
    In this context, previous work~\cite{cooper2016linear} computed the probabilities of each colour winning the process, also known as the fixation probability of a given colour. They show that such probabilities depend on the stationary distribution, $\mu$, of the out-matrix of the graph $G$. To better understand the effect of each node within a graph, we will denote $\mu(v)$ as the \textdef{influence of a vertex $v$}. Observe that the out-matrix $H$ of the graph $G$ can be seen as the transition matrix of a time homogeneous Markov chain (e.g., see Chapter $6$, \citenptext{grimmett2001probability})  representing the probabilities of one round in the consensus process~\cite{cooper2016linear}.
    
    If $G$ is strongly connected, this Markov chain is irreducible and finite, so there exists a unique stationary distribution $\mu$ of $H$, that is, there is a row vector $\mu$ such that $\mu H=\mu$. We call the values $\mu(v)$ the influence of the vertex $v$ in the consensus protocol. The winning probabilities of each colour can be determined by the initial configuration only and are given by the following proposition.
    
    \begin{proposition}[\citenp{cooper2016linear}]\label{prop:nicola's-result}
        Consider a consensus process on a well-behaved (and strongly connected) graph $G$ (i.e., with finite consensus time for all initial configurations), with associated adjacency matrix $H$ and $\mu$ its unique stationary distribution. Assume the initial configuration is given by $\state\in\{c_1,\dots,c_k\}^{V}$. Then, we have that the winning probability of colour $c_i$ is: 
        \[\mathrm{P}(\textrm{colour }c_i\textrm{ wins\textbar}\State_0 = \state) = \sum_{v \in V, \State(v)=c_i}\mu(v)\]
    \end{proposition}
    
    \begin{example}\label{exm:prob}
        Consider the initial configuration discussed in Example \ref{exm:memoryless}. The adjacency matrix of this example is given by
            \[\begin{pmat}({})
                    \,0\, & \,1\, & \,0\, \cr
                    \frac{1}{4} & 0 & \frac{3}{4} \cr
                    0 & \frac{1}{3} & \frac{2}{3} \cr
            \end{pmat}\]
            and its stationary distribution is
            $\mu=\left(\frac{1}{14} \;\;\; \frac{4}{14} \;\;\; \frac{9}{14}\right).$

        Let $\state$ be the initial configuration depicted in Figure~\ref{fig:example_1}. The graph $G$ is {\well}, as can be immediately concluded from the fact that it contains a loop, so Proposition \ref{prop:nicola's-result} can be applied, and thus the winning probabilities are:
                $P(\mbox{{\blue} wins}|\State_0=\state)=\mu(v_1)+\mu(v_2)=\frac{5}{14}$, and 
            $P(\mbox{{\red} wins}|\State_0=\state)=\mu(v_3)=\frac{9}{14}.$

        Note that although the number of {\red} vertices in this initial configuration is smaller than the number of {\blue} vertices, the influence of the vertex $v_3$ is much higher than the influence of $v_1$ and $v_2$. For this reason, {\red} has a higher probability of winning the process.
    \end{example}

    \subsection{Memory Consensus Protocol}
    
    We now introduce the main concept to be explored in this work. The main difference of the process with memory is that each node may take into account the previous states of its neighbours. The notion of consensus also needs to be updated.

    \begin{definition}[Memory Consensus Protocol]\label{def:memory-protocol}
        A $m$-memory consensus protocol generalises the notion of memoryless protocol by changing the rule with which nodes update their colour. At each round $t$ and for $0\leq i\leq m$, each node $v$ chooses a time $t-i$, with probability $p_i$, and copies the colour of one of its out-neighbours proportionally to the weight of the edge in $G$. Given the initial states $\memState_i = \state_i$ for $0 \leq i \leq m$, let $\memstochastic$ be a random variable that records the colours of the set of nodes of a graph $G$ at time $t$. The configuration $\memState_{t+1}$ therefore depends on the states $\memState_t,\dots, \memState_{t-m}$. We call this a memory consensus process $(p_0,\dots,p_m)$ on $G$.

    \end{definition}
    The notion of consensus in memory process needs to be updated when compared to the memoryless one, since a process may move away from a consensus in a current round by agents remembering past states. 
    We then say a memory consensus process reached \textdef{consensus} if and only if it reached a stable consensus, i.e., all nodes have the same colour \textbf{and} there is no positive probability that a node changes colour in any following round.
    
    In Definition \ref{def:memory-protocol}, we have assumed that the initial $(m+1)$-states are fixed arbitrarily, thus enabling the $m$-memory consensus processes to be considered from that state onwards. There might be situations, however, where such records are not available, for example, when a process has just started. For that reason we need a convention on how the memory will be built up. In this work, we will set the convention that if less than $(m+1)$-states are known, then the unknown states are treated as if they are all equal to the first (`oldest') state $\memState_0$ and we act as if we had $m+1$ states in memory. Throughout this paper, we will look closer at memory processes which start with only one given initial state, as we formally define below.

    \begin{definition}[Early Memory Process]\label{def:early-memory-protocol}
        We define the \textdef{early memory process $(p_0,\dots,p_m)$ on $G$ starting at $s\in \Colour^V$} as the memory process $(p_0,\dots,p_m)$ on $G$ with initial configurations $\memState_0=\dots=\memState_m=\state$.
    \end{definition}

\section{Framework and Theoretical Results for Winning Probabilities}\label{sec:convergence-probability}
    Analysing processes with memory can be hard since standard Markov chain tools, such as the ones described in Proposition \ref{prop:nicola's-result}, cannot be applied. In this section, we propose a framework to study processes with memory by creating an equivalent process which is itself memoryless. 
    
    We use this framework to study whether the deadlocks discussed in the previous section can be avoided
    in memory processes. Further we study the probabilities of each colour to win an ongoing $m$-memory process taking into account the current round together with the previous $m$ configurations.
    Finally, we compare the probabilities of consensus of a given colour between the memoryless and early memory settings.

    We summarise the discussion above as a sequence of four questions to be explored in this section. 
    \begin{questions}
        \item\label{itm:reduction} \textbf{Reduction to Memoryless Case:} Can we reduce a memory consensus process to a memoryless one in order to make use of previously known results? 
        \item\label{itm:no-losing-configs} \textbf{Well-Behaved Graphs:} Can deadlocks arise in memory processes?
        \item\label{itm:who-wins} \textbf{Who Wins:} Given an arbitrary memory consensus process, what
        is the probability of each colour winning?
        \item\label{itm:inertia} \textbf{Memory vs Memoryless Processes:} Given the same initial state for both an early memory and a memoryless consensus process, how do winning probabilities 
        compare?
        \setcounter{questionscounter}{\value{enumi}}
    \end{questions}

    \subsection{A Framework to Study Memory Processes}
        We start by addressing Question \ref{itm:reduction}. The intuition is to transform a non-Markovian process (due to dependency on several previous states), into a Markovian one. From a graph $G$, we create a new graph, $\mG$, that captures the previous $\maxmemory$ rounds of a process on $G$. For that, $\mG$ contains $\maxmemory$ extra copies of the set of nodes of $G$ (each copy is called a layer of $\mG$), to represent past $\maxmemory$ configurations of $G$.  Edges are added to simulate desired behaviour of nodes: nodes representing past states simply copy the state of their `future', i.e., nodes that are one layer above. Moreover, edges leaving nodes that represent the present may `access' the information stored in the layers below. A formal definition of a memory graph is given below. 
        
        \begin{definition}[$m$-Memory Graph]\label{def:memory-graph} Let $G=(V,E)$ be a directed weighted graph, with $V=\{v_1,\dots,v_n\}$. For each $m\geq 0$ and $p_0,\dots,p_m>0$ with $\sum p_i=1$, we define a directed weighted graph $\mG=(\mV,\mE)$ called the \textdef{associated $m$-memory graph with probabilities $(p_0,\dots,p_m)$}. The set $\mV$ is given by \[\{v_{ij}|\text{ $i=0,\dots,m$ and $j=1,\dots n$}\},\] and we say that the collection $\{v_{ij}:j=1,\dots,n\}$ is the \textdef{$i$th layer of the graph} $\mG$. The edges in $\mE$ are of three types:
            \begin{itemize}
                \item Horizontal edges: if $(v_j,v_k)\in E$ with weight $w$ then $(v_{0j},v_{0k})\in \mE$ with weight $p_0 w$.
                \item Descending edges: if $(v_j,v_k)\in E$ with weight $w$ then $(v_{0j},v_{ik})\in \mE$ with weight $p_i w$, for all $i >0$.
                \item Ascending edges: for every $i>0$ and $j$, there is an edge from $v_{ij}$ to $v_{(i-1)j}$ with weight $1$;
            \end{itemize}
        \end{definition}
        
        Note that, by definition, the only $0$-memory graph associated to a graph $G$ is $G$ itself. To help understanding graph $\mG$, we presents its adjacency matrix. Let $H$ denote the adjacency matrix of the graph $G$. Then it follows from the definition that the adjacency matrix of the associated $m$-memory graph with probabilities $(p_0,\dots,p_m)$ is given by
            \begin{equation}\label{eq:matrix_stationary_distribution_memory}
                \overline{H} =\begingroup
                    \pmatset{3}{14pt}
                    \pmatset{4}{16pt}
                    \pmatset{5}{14pt}
                    \pmatset{6}{5pt}
                    \begin{pmat}({|||})
                        H_0 & H_{1} & \dots & H_{m} \cr\-
                        I & 0 & 0 & 0 \cr\-
                        0 & \ddots & \ddots & 0 \cr\-
                        0 & 0 & I & 0 \cr
                    \end{pmat}
                \endgroup
            \end{equation}
        where each $H_i=p_i H$ for $i=0,\dots,m$, and the vertices of $\mG$ are ordered $v_{01},v_{02},\dots,v_{mn}$.
        
        Now, we motivate the definition of the associated memory graph $\mG$ of $G$, by showing that a (memoryless) consensus process in $\mG$ a the memory consensus process $(p_0,\dots, p_m)$ on $G$ are equivalent. To define this equivalence precisely, we need the following definition.

        \begin{definition}\label{def:associated-memoryless}
            Let $\{\memState_0,\dots,\memState_m\}$ be the first $m$ rounds of a memory consensus process $(p_0,\dots,p_m)$ on a graph $G$. We define $\mstochastic$ the \textdef{associated memoryless consensus process} as the process on $\mG$, the associated $m$-memory graph with probabilities $(p_0,\dots,p_m)$. The initial state $\MState_m(v_{ij})$ of a vertex $v_{ij}$ in the $i$th layer of $\mG$ is defined to be $\memState_{m-i}(v_j)$.
        \end{definition}

        We can now show an example of a memory consensus process on a memory graph by extending our original graph example from Figure \ref{fig:example_1}.
        
        \begin{example}\label{exm:memorygraph}
            If $G$ is the graph considered in Example~\ref{exm:memoryless}, then 
            Figure~\ref{fig:ex-mem-2} shows the $2$-memory graph with probabilities $(\frac{1}{3},\frac{1}{3},\frac{1}{3})$ associated to $G$ in which colours refer to a possible initial configuration of the process on $\mG$.

            \begin{figure}
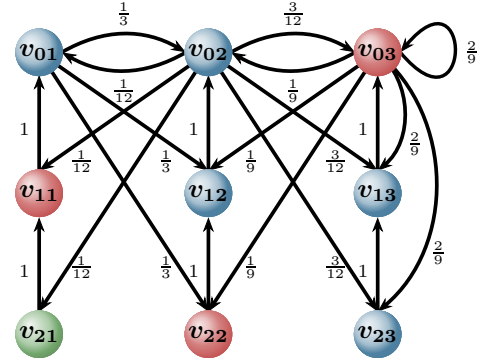

                \centering
                \twomemsamplegraph
                \caption{A Possible Initial Configuration of a Consensus Process on the $2$-Memory Graph $\mG$ with Probabilities $(\frac{1}{3},\frac{1}{3},\frac{1}{3})$ Associated to $G$.}\label{fig:ex-mem-2}
            \end{figure}

        \end{example}
        
        Settling Question \ref{itm:reduction}, the following proposition shows that $\mstochastic$ from Definition \ref{def:associated-memoryless} is indeed the memoryless equivalent to our memory protocol from Definition \ref{def:memory-protocol}. The following proposition shows that the two processes have the same distribution at every round $t\geq m$.

        \begin{proposition}\label{prop:reduction}
            Let $\memstochastic$ be a memory consensus process $(p_0,\dots,p_m)$ on a graph $G$ and let $\mstochastic$ be the associated memoryless consensus process on $\mG$. Let $\memstate_i$ be a configuration on $G$, for $i=0,\dots,m$. Then, for any $\round\geq m$: 
                \[P(\cup_{i=0}^m(\memState_{\round-i}=\memstate_i)|\memState_0,\dots,\memState_{m}) = P(\MState_\round=\bar{s}|\MState_m),\]
            where $\bar{s}$ is the configuration on $\mG$ where layer $i$ has configuration $s_i$.
        \end{proposition}
        \begin{proof}
            We proceed by induction on $\round$. For $\round=m$, it is trivial matter as both distributions are deterministic. Assuming the induction hypothesis, i.e.,  that for $\round=\round_0$ it holds that:
                \[P(\cup_{i=0}^m(\memState_{\round_0-i}=\memstate_i)|\memState_0,\dots,\memState_{m})=
                P(\MState_{\round_0}=\bar{s}|\MState_m)\]
            for every fixed configuration $\memstate_i$, for $i=0,\dots, m$. We now prove the induction step from $\round_0$ to $\round_0+1$. By Definition \ref{def:memory-protocol} we know that, for every fixed configuration $\memstate_i$, for $i=0,\ldots, m$:
                \[\begin{aligned}
                     P(\cup_{i=0}^m(&\memState_{\round_0+1-i}=\memstate_i)|\memState_0,\dots,\memState_{m}) =\\= \textstyle\sum_{r_i} \big[& P(\cup_{i=0}^m(\memState_{\round_0-i}=r_i)|\memState_0,\dots,\memState_{m})\cdot\\& P(\cup_{i=0}^m(\memState_{\round_0+1-i}=s_i)|\cup_{i=0}^m(\memState_{\round_0-i}=r_i)) \big]
                \end{aligned}\]
            where the sum is over all possible choices of configurations $r_i$. Note that if $\bar{r}$ is such that layer $i$ of $\mG$ receives configuration equal to $r_i$ then we have that:
                \[\begin{aligned}
                    P(\cup_{i=0}^m(\memState_{\round_0-i}=r_i)|\memState_0,\dots,\memState_{m}) = P(\MState_{\round_0}=\bar{r}|\MState_m)
                \end{aligned}\]
            by the induction hypothesis. Moreover, inspecting the adjacency matrix of $\mG$ and $G$ it is trivial to conclude that
            $P(\cup_{i=0}^m(\memState_{\round_0+1-i}=s_i)|\cup_{i=0}^m(\memState_{\round_0-i}=r_i))$ is equal to $P(\MState_{\round_0}=\bar{s}|\bar{r})$.
            Combining the two we obtain the desired result, by noting that we have 

                    $P(\MState_{\round_0+1}=\bar{s}|\MState_m) = \sum_{\bar{r}}P(\MState_{\round_0}=\bar{r}|\MState_m)P(\MState_{\round_0+1}=\bar{s}|\MState_{\round_0}=\bar{r}).$
        \end{proof}

    \begin{corollary}[Convergence Times and Probabilities]
        A memory process $\memstochastic$ on $G$ and a process $\mstochastic$ on $\mG$, under the conditions of Proposition \ref{prop:reduction}, have the same expected number of rounds until consensus, and the same probability of convergence for each colour $c \in \Colour$.
    \end{corollary}

    We now have all the tools necessary to analyse the memory consensus protocol while addressing Questions  \ref{itm:no-losing-configs} to \ref{itm:inertia}. This is done in the section that follows.

\subsection{Results on Probabilities of Consensus}

The concept of the memory graph associated to a memory process allows us to translate standard results about memoryless consensus processes to this new context. In this section, we show how one can use standard results to discuss convergence of memory processes and their probabilities of consensus for each colour. 

We start by answering Question~\ref{itm:no-losing-configs} by showing that $m$-memory consensus processes always converge (as long as, of course, $m >0$). This is the first key benefit arising from the  memory protocol compared to their memoryless counterpart. In particular, memory processes in all graphs discussed in Example \ref{exm:badly-behaved} now reach consensus with probability $1$.

\begin{proposition}[Memory Graphs are Well-Behaved]\label{prop:no-losing-configs}
    Let $\mG$ be a $m$-memory graph with $m >0$ associated to a memory process in a (strongly connected) graph $G$. Then, $\mG$ is {\well}. 
\end{proposition}
\begin{proof}
    The proof is a consequence of Proposition~\ref{prop:gcd}. Consider a cycle in $G$, which exists because $G$ is strongly connected. Denote the cycle by $(v_1, v_2,... v_k)$. Then, $\mG$ contains the cycle $(v_{01}, v_{02}, ...v_{0k})$ of length $k$. Moreover, $\mG$ also contains the cycle $(v_{01}, v_{12}, v_{02}, ...v_{0k})$, of length $k +1$. Therefore we have cycles in $\mG$ of lengths $k$ and $k+1$, which implies that the gcd of the length of all cycles is $1$. 
\end{proof}

To settle Question~\ref{itm:who-wins}, we determine the probability of consensus for $m$-memory processes. That is, given the current and also the previous $m$ rounds of a $m$-memory consensus process, we give exact probabilities of each colour winning.

\begin{theorem}\label{thm:winning-probabilities}
    Let $\memstochastic$ be a memory consensus process $(p_0,\dots,p_m)$ on a (strongly connected) graph $G$ with initial states $\memState_i=\memstate_i$, for $i=0,\dots,m$. Let $\mu$ be the stationary distribution of $G$. Then for any colour $c\in \Colour$, the winning probability is given by
        \[\begin{aligned}
            \pmem(c \,& \text{wins}|\memState_0=\memstate_0,\dots,\memState_m=\memstate_m)=\\&=\sum\limits_{i=0}^m\frac{1-p_0-\dots-p_{i-1}}{\sigma}\left(\sum\limits_{\memstate_{m-i}(v_{j})=c}\inf(v_{j})\right)
        \end{aligned}\]
    where $\sigma=p_0+2p_1+3p_2+\dots+(n+1)p_n$.
\end{theorem}

    \begin{proof}
        By Proposition~\ref{prop:reduction}, the probability \[\pmem(c \, wins|\memState_0=\memstate_0,\dots,\memState_m=\memstate_m)\] is equal to the probability of $c$ winning the associated memoryless process on $\mG$, the associated $m$-memory graph with probabilities $(p_0,\dots,p_m)$ and initial configuration $\mstate$, as described in Definition~\ref{def:associated-memoryless}. We will calculate this probability using Proposition~\ref{prop:nicola's-result}.
        
        Recall that the adjacency matrix of the process $\mG$ is given by~\eqref{eq:matrix_stationary_distribution_memory}. Let $\mu$ be the stationary distribution of the graph $G$. Using that $H_{i}=p_i H$ and $\mu H=\mu$, it is easy to check that the stationary distribution of $\mG$ is given by 
            \begin{equation}\label{eq:weight-of-layers}
                \msdis=\frac{1}{\sigma}(v, \;\; \alpha_1 v, \;\; \alpha_2 v, \;\; \dots\;\;, \alpha_n  v)
            \end{equation}
        where $\alpha_i=1-p_0-\dots-p_{i-1}$ and $\sigma=p_0+2p_1+3p_2+\dots+(n+1)p_n$.
        
        Then by Proposition~\ref{prop:nicola's-result} the probability of colour $c$ winning the memoryless process on $\mG$ with initial configuration $\mstate$ is
            \[\begin{aligned}
                &\pmem(c \, \text{wins on }\mG|\MState_m=\mstate)= \\
                    &=\textstyle\sum\limits_{v\in \mV, \mstate(v)=c}\msdis(v)
                    =\textstyle\sum\limits_{i=0}^m\textstyle\sum\limits_{\mstate(v_{ij})=c}\msdis(v_{ij})\\
                    &=\textstyle\sum\limits_{i=0}^m\textstyle\sum\limits_{\mstate(v_{ij})=c}\frac{\alpha_i}{\sigma}\cdot\inf(v_{j})
                    =\textstyle\sum\limits_{i=0}^m\frac{\alpha_i}{\sigma}\left(\textstyle\sum\limits_{\state_{m-i}(v_{j})=c}\inf(v_{j})\right)
            \end{aligned}\]
    \end{proof}

     \begin{example}\label{exm:prob-memory}
        Let Figure~\ref{fig:ex-mem-2} be the initial configuration $\mstate$ on the memory graph $\mG$ associated to a memory consensus process $(\frac{1}{3},\frac{1}{3},\frac{1}{3})$ on the graph $G$ of Figure~\ref{fig:example_1}. Using the stationary distribution for $G$ which was computed on Example~\ref{exm:prob} and Theorem~\ref{thm:winning-probabilities}, we get that the stationary distribution of $\mG$ is given by the vector
            \[\begin{aligned}
                \msdis=\frac{1}{2}\left(\frac{1}{14} \;\;\; \frac{4}{14} \;\;\; \frac{9}{14} \;\;\; \frac{1}{21} \;\;\; \frac{4}{21} \;\;\; \frac{9}{21} \;\;\; \frac{1}{42} \;\;\; \frac{4}{42} \;\;\; \frac{9}{42}\right).
            \end{aligned}\]
        The probabilities of consensus are given by Proposition
         \ref{prop:nicola's-result}:
            \[\begin{aligned}
                P(&\mbox{{\blue} wins}|\MState_0=\mstate)=\\
                    &=\frac{1}{2}(\inf(v_{1})+\inf(v_{2})) + \frac{1}{3}(\inf(v_{2})+\inf(v_{3}))+ \frac{1}{6}\cdot\inf(v_{3})\\
                    &=\frac{1}{2}\cdot\frac{5}{14} + \frac{1}{3}\cdot\frac{13}{14}+ \frac{1}{6}\cdot\frac{9}{14}=\frac{25}{42}\\
            \end{aligned}\]
            Analogously, we conclude that $P(\mbox{{\red} wins}|\MState_0=\mstate)= \frac{11}{28}$ and $P(\mbox{{\green} wins}|\MState_0=\mstate)=\frac{1}{84}$.

    \end{example}

    In the example just presented, the past rounds had an effect on which colour is more likely to win the memory process. But what exactly is the influence of the past? In other words, what is the combined influence of nodes in a layer compared another? The answer to that was given in Equation \eqref{eq:weight-of-layers}. For the scenario in Example \ref{exm:prob-memory}, the combined influence of nodes in layer $0$ is $\frac{1}{2}$, layer $1$ is $\frac{1}{3}$, and  layer $2$ is $\frac{1}{6}$. Note that influences are in descending order. Indeed, for any $m$-memory process, layer $i$ always has more influence than layer $j$ for $j > i$ for any values $p_0, \dots, p_m$.

    Finally, we compare memoryless processes with their correspondent early memory version with regards to probabilities of convergence (Question \ref{itm:inertia}).

\begin{corollary}\label{cor:memory-maintains-probability}
    An early $m$-memory process and a memoryless process starting at the same initial configuration on a well-behaved graph have the same probabilities of convergence for each colour. 
\end{corollary}

\begin{proof}
    Let $\pnomem(c \text{ wins on }G|\State_0=\state)$ be the probability of $c$ winning the memoryless process on $G$. Consider an early memory process $(p_0,\dots,p_m)$ on $G$ with starting configuration $\State_0=\state$, for any choice of $p_0,\dots,p_m$. By Definition~\ref{def:early-memory-protocol}, we know $\memState_{m-i}(v_{j})=c$ if and only if $\state(v_{j})=c$. Let $\alpha_i=1-p_0-\dots-p_{i-1}$, then applying the result of Theorem \ref{thm:winning-probabilities}, the probability of $c$ winning is 
        \[\begin{aligned}
            \pmem&(c \, \text{wins on }\mG|\MState_m=\mstate) 
            =\left(\textstyle\sum\limits_{\state(v_{j})=c}\inf(v_{j})\right)\left(\textstyle\sum\limits_{i=0}^m\frac{\alpha_i}{\sigma}\right)=\\
            &=\textstyle\sum\limits_{\state(v_{j})=c}\inf(v_{j})=\pnomem(c \text{ wins on }G|\State_0=\state)\\
        \end{aligned}\]
    where the second to last equality follows from the fact that $\sum\limits_{i=0}^m \alpha_i=\sum\limits_{i=0}^m(p_i+{p_{i+1}}+\dots+p_{n})=\sigma$.
\end{proof}

An equivalent result to Corollary~\ref{cor:memory-maintains-probability} is that the influence of a node in a memoryless process on $G$ is the same as the sum of influences of this same node and its $m$ copies on $\mG$. This is an advantage of memory when compared to the strategy of avoiding deadlocks on memoryless processes by including new edges, as the latter may change the influence of the nodes in the process.

\subsubsection{A Note on a More General Memory Protocol}
For readability, motivation, and presentation, we have introduced a memory protocol assuming all agents have the same probabilities of  remembering past rounds, and not allowing agents to remember nodes  that they are not connected to in the present. To lift these assumptions is to consider a memory graph in which nodes representing the present may be arbitrarily linked with past layers, as long as the weight of edges adds up to $1$. The probabilities of consensus of this framework can be established using Proposition \ref{prop:nicola's-result}, as long as the graph is {\well}. If not, techniques from~\citenptext{marzagao2017team} can be used to apply analogous results for any $m$-memory process on arbitrary directed graphs $G$.

\section{Empirical Analysis of Duration of Processes}\label{sec:empirical-analysis}

In this section we investigate, through simulations, how the duration (measured in number of round until consensus) of early $1$-memory process compares to their memoryless counterparts. We will restrict ourselves to processes on undirected graphs, a set $\Colour = \{\red, \blue\}$, and only one layer of memory ($m=1$). 
When considering the $1$-memory consensus process $(p_0,p_1)$ on undirected graph $G$, we assume that, with probability $p_0$ (resp. $p_1$), a node copies the present (resp. past) colour of a neighbour chosen uniformly at random.
The investigation of duration of $m$-memory processes for $m \geq 2$ is subject to future work.

Recall that the condition for a memory process to reach a (stable) consensus is stronger than for memoryless ones: in a $m$-memory process, we not only need all nodes to have the same state in the present round, but also in all the previous $m$ rounds, so there is no chance that an agent changes colour based on a past state of a neighbour.

We have chosen different standard network structures to analyse: cliques (complete graphs), cycles, 
bicliques (complete bipartite graphs with an extra loop at the larger size), full binary trees (with a loop at the root), and grids on a torus (two dimensional grids with connected ends). We have added loops to the 
full binary tree and to the biclique because we need the graphs to be {\well} (otherwise, they may never reach a consensus). This selection offers a wide range of graph densities, as well varying averages for consensus times in memoryless processes, as will be discussed shortly.

\begin{table}[]
    \centering
    \begin{tabular}{c|c|c}
        Topology & Average Consensus Time & Median \\ \hline \hline
        clique & $1,420 \pm 1,049$ & $1,127$ \\ \hline
        grid & $3,711 \pm 2,827$ & $2,943$\\ \hline
        bintree & $15,033 \pm 11,589$ & $11,681$\\ \hline
        biclique & $131,939 \pm 222,579$ & $3,060$\\ \hline
        cycle & $438,232 \pm 434,180$ & $305,546.5$ \\ \hline
    \end{tabular}
    \caption{Average, standard deviation, and median for memoryless consensus times on graphs of size $n=1023$ over $4000$ runs. }
    \label{tab:times-memoryless}
\end{table}

We perform two experiments on the network topologies described above. Each experiment compares a memoryless process with a given initial configuration with its early $1$-memory counterpart with the same initial state, similarly to what was discussed in the context of Question \ref{itm:inertia}. To avoid bias given by the initial state, each set of experiments (with and without memory) has a different random starting point, with each node being $\red$ or $\blue$ with equal probability. For a given pair $(p_0, n)$ and a graph type, we denote the  \textdef{the ratio between the average consensus time of the $1$-memory process and the average consensus time of its memoryless counterpart} by $\average$. Thus, $\average > 1$ (resp. $\average < 1$) indicates memory processes take longer (resp. shorter) than memoryless ones. 
In the first experiment, we fix the number of nodes $n$,  while varying $p_0$ to explore the effect of memory for these different values. The second experiment fixes a value of $p_0$ to investigate how improvement of memory changes with $n$.

In experiment $1$, we have recorded the duration of $4000$ simulations for graphs of size $n=1023$, for $30$ different values of $p_0$, ranging uniformly from $0.1$ to $1$. The value $n=1023$ was chosen to allow for binary trees to be full and the torus to have similar dimensions ($31$ and $33$). Table \ref{tab:times-memoryless} shows the average times for consensus in the memoryless case (i.e., $p_0 = 1$) as well as the standard deviation and median. Note that in consensus processes standard deviations are particularly high, of the order of magnitude of the average itself. 
For that reason, we calculate the median for each graph type, showing that it is well below the average. The full data, including the process duration distributions, all data points, and analogous plots with the median instead the average (which show less pronounced but similar results) can be found at \url{https://github.com/tmadeira/consensus}. 

\begin{figure}
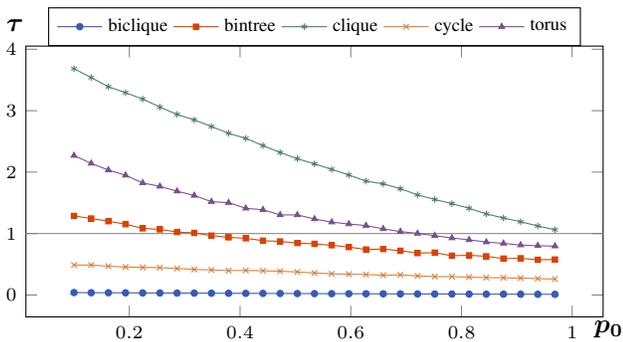

        \centering
        \consensustimefixedn
        \caption{[Experiment 1] A comparison of $1$-memory processes and their memoryless counterparts, i.e., $\average$ values (y-axis) for a fixed $n = 1023$ and different values of $p_0$ (x-axis) and five network topologies. 
        }
        \label{plot:memory-cycle-improvement}
    \end{figure}

Results of Experiment 1 are shown in Figure \ref{plot:memory-cycle-improvement} with $x$-axis indicating the different values of $p_0$, whereas the values on the $y$-axis represent $\average$. Note that taking $p_0=1$ is the same as having a memoryless process, 
so we have omitted this value from the graph in Figure \ref{plot:memory-cycle-improvement}.

For all values of $p_0$, there is a considerable improvement in the average consensus time for memory processes on the cycle and biclique, the latter being the type that benefits the most from memory, irrespective of $p_0$, with ratios ranging from $0.01$ (for $p_0 = 0.97)$ to $0.04$ (for $p_0 = 0.1$). For processes on a grid and binary trees, there is no gain for small values of $p_0$, but for larger values of $p_0$, the consensus times on the torus is improved in the presence of memory. 

Based on the results of Figure~\ref{plot:memory-cycle-improvement}, we conjecture that graphs which are in some sense close to bipartite are those which benefit from memory. A precise definition of `closeness' to bipartite graphs is subject to future work.
The intuition, however, is that in memoryless processes on graphs close to bipartite graphs (the biclique with an extra loop being the most extreme example), the partitions behave almost independently: if there are more $\red$ nodes in a given partition and more $\blue$ nodes in the other, then it becomes very likely that this picture will be inverted in the following round. With the addition of memory, on the other hand, this vicious cycle can more easily be broken, thus decreasing the average consensus times. The median being substantially lower than the average for bicliques further supports the hypothesis above: whenever partitions tend to the same colour, consensus is very quick. When they do not, however, it may take several orders of magnitude longer.

We now turn to Experiment $2$ that looks at how $\tau$ changes when $n$ varies. To do this, we fix the probability $p_0=0.9$. Our choice is motivated by the result of previous experiment that indicated $0.9$ is among the best values for improving the average convergence time when compared to memoryless processes. The setup is analogous to the one in Experiment $1$, with the difference that we now average over $10^4$ simulations for each $n$ and each graph type. The values chosen for $n$ depend on the type of graph. The number of nodes on a full binary tree is always $2
^k - 1$, so we used for our test all such values for $k \in \{3, 11\}$. On the other hand, the number of vertices for a {\well} square grid on a {\torus} needs to be a perfect square of an odd number. So for testing all other types of network structures, we used all such numbers from $9$ to $2025$.

\begin{figure}
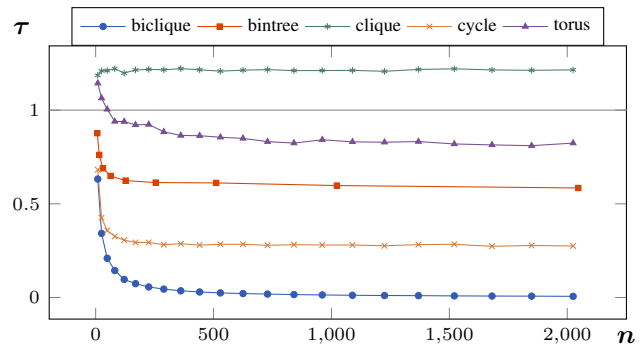

        \centering
        \consensustimefixedp
        \caption{
        [Experiment 2] A comparison of $1$-memory processes and their memoryless counterparts, i.e., $\average$ values (y-axis) for fixed $p_0 =0.9$ and different values of $n$ (x-axis) and five network topologies. 
        }
        \label{plot:fig2}
    \end{figure}

The results are shown in Figure \ref{plot:fig2} with $x$-axis indicating the number of nodes $n$, whereas the values on the $y$-axis represent the ratio $\average$. We can see from this experiment that, for all graph types but cliques, the benefit of memory increases as $n$ increases, but soon stabilises for $n \approx 800$. This supports the claim that improvements in convergence times given by memory are not a feature only of small graph sizes. To show robustness of improvement from the use of memory, we performed a two-sample t-test statistic for means and rejected a null hypotheses of no difference in means between the memory and memoryless processes with $>99.99\%$ confidence. We considered all graph classes, apart from clique, and chose $n$ as the lowest value among the ones tested that was greater than $1000$ in each class.

\section{Related Work}\label{sec:related-work}

Memoryless consensus protocol is also known as voter model and have been extensively studied in the literature~\cite{donnelly1983finite,nakata1999probabilistic, hassin2001distributed, aldous1995reversible}. 
Linear voting model, described in~\cite{cooper2016linear}, are a generalization of this process. Those encompass all typically studied forms of voting models such as push or pull models. Previous work has characterised graphs for which this process converges almost surely~\cite{marzagao2017team}, computed the winning probabilities for each colour and given bounds on the convergence time~\cite{oliveira2012coalescence, oliveira2013mean, cooper2016linear, marzagao2017multi, kanade2019coalescence, oliveira2019random}. 

In the context of control theory, protocols where agents remember their past states have been previously studied~\cite{cao2008multi,li2010distributed}. In the context of multi-agent networks, protocols where agents have memory were also explored~\cite{pasolini2020exploiting}. Unlike ours, such protocols are in a continuous setting. In the context of vehicle coordination, they model each vertex with a certain position and associated speed. In the context of multi-agent networks, a scalar measurement is propagated so that the mean measurement is computed. We note that the authors find a similar result to the one here, namely, the addition of memory speeds up the convergence of their agents. There is also a generalization of the voter model \cite{zhong2016generalized} which considers agents with memory of past states. In their protocol and topologies studied, the authors find memory to be detrimental to consensus.
When fault of systems are considered, authors in \cite{continuous-consensus} studied processes known as `continuous consensus'. In these, nodes may keep  information about the  past in order for all processor to reach a common value. 

\section{Conclusions and Future Work}\label{sec:conclusions}

This paper introduced a generalisation of synchronous consensus protocols, by the addition of memory of previous states influencing the decisions of nodes at each round. Probabilities of consensus for each starting configuration are computed by using previously known results and it is shown empirically that memory is beneficial to convergence on typical network topologies, such as  cycles.

Future work may explore theoretical convergence times in different graph structures when memory is introduced. Note that, while general upper bounds from previous work (e.g.~\cite{cooper2016linear}) apply to the memory case via the memory graph, these bounds are too loose and do not even exhibit the observed qualitative behaviour that cycles, say, benefit from memory. On the experimental side, it is key to establish whether there exist graphs in which $2$-memory processes converge faster on average than do $1$-memory processes. A discussion of the tradeoff between adding layers and gain in speed is also pertinent for the use of memory framework in realistic settings. Although we only analyse a consensus protocol similar to the voter model, our framework (Definition \ref{def:memory-graph}) may be used to analyse other models such as majority rule \cite{mossel2014majority} or average consensus \cite{tsitsiklis1986distributed}.  
Another interesting issue would be to characterise which graph structures benefit from the addition of memory and which do not. Lastly, while most of our theoretical results apply to the case of three or more colours, a more in-depth empirical analysis of consensus times in this more general case would be interesting.

\section{Code and Data}\label{sec:code-and-data}
The repository containing the code, data, and plots associated to this project can be found at \url{https://github.com/tmadeira/consensus}. 
        
\section{Acknowledgements}
The work by David Kohan Marzagão relates to Department of Navy award (Award No. N62909-18-1-2079) issued by the Office of Naval Research. The United States Government has a royalty-free license throughout the world in all copyrightable material contained herein.

During the development of this work Luciana Basualdo Bonatto was supported by CNPq (201780/2017-8).

The authors thank Thiago R. Oliveira and Josh Murphy for their helpful comments on earlier versions of this paper.

\bibliography{main.bib}

\begin{thebibliography}{27}
\providecommand{\natexlab}[1]{#1}
\providecommand{\url}[1]{\texttt{#1}}
\providecommand{\urlprefix}{URL }
\expandafter\ifx\csname urlstyle\endcsname\relax
  \providecommand{\doi}[1]{doi:\discretionary{}{}{}#1}\else
  \providecommand{\doi}{doi:\discretionary{}{}{}\begingroup
  \urlstyle{rm}\Url}\fi

\bibitem[{Aldous and Fill(1995)}]{aldous1995reversible}
Aldous, D.; and Fill, J. 1995.
\newblock Reversible Markov chains and random walks on graphs.
\newblock Monograph. Berkeley, CA, USA.

\bibitem[{Cao, Xiao, and Wang(2015)}]{cao2015event}
Cao, M.; Xiao, F.; and Wang, L. 2015.
\newblock Event-based second-order consensus control for multi-agent systems
  via synchronous periodic event detection.
\newblock \emph{IEEE Transactions on Automatic Control} 60(9): 2452--2457.

\bibitem[{Cao, Ren, and Chen(2008)}]{cao2008multi}
Cao, Y.; Ren, W.; and Chen, Y. 2008.
\newblock Multi-agent consensus using both current and outdated states.
\newblock \emph{IFAC Proceedings Volumes} 41(2): 2874--2879.

\bibitem[{Cooper and Rivera(2016)}]{cooper2016linear}
Cooper, C.; and Rivera, N. 2016.
\newblock The linear voting model.
\newblock In \emph{43rd International Colloquium on Automata, Languages, and
  Programming (ICALP 2016)}. Schloss Dagstuhl-Leibniz-Zentrum fuer Informatik.

\bibitem[{Donnelly and Welsh(1983)}]{donnelly1983finite}
Donnelly, P.; and Welsh, D. 1983.
\newblock Finite particle systems and infection models.
\newblock In \emph{Mathematical Proceedings of the Cambridge Philosophical
  Society}, volume~94, 167--182. Cambridge University Press.

\bibitem[{Grimmett et~al.(2001)Grimmett, Grimmett, Stirzaker
  et~al.}]{grimmett2001probability}
Grimmett, G.; Grimmett, G.~R.; Stirzaker, D.; et~al. 2001.
\newblock \emph{Probability and Random Processes}.
\newblock Oxford University Press.

\bibitem[{Hassin and Peleg(2001)}]{hassin2001distributed}
Hassin, Y.; and Peleg, D. 2001.
\newblock Distributed probabilistic polling and applications to proportionate
  agreement.
\newblock \emph{Information and Computation} 171(2): 248--268.

\bibitem[{Ismail and Sariff(2018)}]{ismail2018survey}
Ismail, Z.~H.; and Sariff, N. 2018.
\newblock A survey and analysis of cooperative multi-agent robot systems:
  challenges and directions.
\newblock In \emph{Applications of Mobile Robots}. IntechOpen.

\bibitem[{Kanade, Mallmann-Trenn, and Sauerwald(2019)}]{kanade2019coalescence}
Kanade, V.; Mallmann-Trenn, F.; and Sauerwald, T. 2019.
\newblock On coalescence time in graphs: When is coalescing as fast as meeting?
\newblock In \emph{Proceedings of the Thirtieth Annual ACM-SIAM Symposium on
  Discrete Algorithms}, 956--965. SIAM.

\bibitem[{Kohan~Marzag{\~a}o et~al.(2017{\natexlab{a}})Kohan~Marzag{\~a}o,
  Murphy, Young, Gauy, Luck, McBurney, and Black}]{marzagao2017team}
Kohan~Marzag{\~a}o, D.; Murphy, J.; Young, A.~P.; Gauy, M.~M.; Luck, M.;
  McBurney, P.; and Black, E. 2017{\natexlab{a}}.
\newblock Team Persuasion.
\newblock In \emph{International Workshop on Theorie and Applications of Formal
  Argumentation}, 159--174. Springer.

\bibitem[{Kohan~Marzag{\~a}o et~al.(2017{\natexlab{b}})Kohan~Marzag{\~a}o,
  Rivera, Cooper, McBurney, and Steinh{\"o}fel}]{marzagao2017multi}
Kohan~Marzag{\~a}o, D.; Rivera, N.; Cooper, C.; McBurney, P.; and
  Steinh{\"o}fel, K. 2017{\natexlab{b}}.
\newblock Multi-agent flag coordination games.
\newblock In \emph{AAMAS}, 1442--1450.

\bibitem[{Li et~al.(2010)Li, Xu, Chu, and Wang}]{li2010distributed}
Li, J.; Xu, S.; Chu, Y.; and Wang, H. 2010.
\newblock Distributed average consensus control in networks of agents using
  outdated states.
\newblock \emph{IET Control Theory \& Applications} 4(5): 746--758.

\bibitem[{Lynch(1996)}]{lynch1996distributed}
Lynch, N.~A. 1996.
\newblock \emph{Distributed algorithms}.
\newblock Elsevier.

\bibitem[{Martinez et~al.(2005)Martinez, Bullo, Cortes, and
  Frazzoli}]{martinez2005synchronous}
Martinez, S.; Bullo, F.; Cortes, J.; and Frazzoli, E. 2005.
\newblock On synchronous robotic networks Part I: Models, tasks and complexity
  notions.
\newblock In \emph{Proceedings of the 44th IEEE Conference on Decision and
  Control}, 2847--2852. IEEE.

\bibitem[{Mizrahi and Moses(2008)}]{continuous-consensus}
Mizrahi, T.; and Moses, Y. 2008.
\newblock Continuous Consensus with Failures and Recoveries.
\newblock volume 5218, 408--422.
\newblock ISBN 978-3-540-87778-3.
\newblock \doi{10.1007/978-3-540-87779-0_28}.

\bibitem[{Mossel, Neeman, and Tamuz(2014)}]{mossel2014majority}
Mossel, E.; Neeman, J.; and Tamuz, O. 2014.
\newblock Majority dynamics and aggregation of information in social networks.
\newblock \emph{Autonomous Agents and Multi-Agent Systems} 28(3): 408--429.

\bibitem[{Nakamoto(2009)}]{nakamoto2008bitcoin}
Nakamoto, S. 2009.
\newblock Bitcoin: A Peer-to-Peer Electronic Cash System.
\newblock \emph{URL \url{http://nakamotoinstitute.org/bitcoin/}.} Last visited
  2021-03-18.

\bibitem[{Nakata, Imahayashi, and Yamashita(1999)}]{nakata1999probabilistic}
Nakata, T.; Imahayashi, H.; and Yamashita, M. 1999.
\newblock Probabilistic local majority voting for the agreement problem on
  finite graphs.
\newblock In \emph{International Computing and Combinatorics Conference},
  330--338. Springer.

\bibitem[{Olfati-Saber, Fax, and Murray(2007)}]{olfati2007consensus}
Olfati-Saber, R.; Fax, J.~A.; and Murray, R.~M. 2007.
\newblock Consensus and cooperation in networked multi-agent systems.
\newblock \emph{Proceedings of the IEEE} 95(1): 215--233.

\bibitem[{Oliveira(2012)}]{oliveira2012coalescence}
Oliveira, R. 2012.
\newblock On the coalescence time of reversible random walks.
\newblock \emph{Transactions of the American Mathematical Society} 364(4):
  2109--2128.

\bibitem[{Oliveira and Peres(2019)}]{oliveira2019random}
Oliveira, R.~I.; and Peres, Y. 2019.
\newblock Random walks on graphs: new bounds on hitting, meeting, coalescing
  and returning.
\newblock In \emph{2019 Proceedings of the Sixteenth Workshop on Analytic
  Algorithmics and Combinatorics (ANALCO)}, 119--126. SIAM.

\bibitem[{Oliveira et~al.(2013)}]{oliveira2013mean}
Oliveira, R.~I.; et~al. 2013.
\newblock Mean field conditions for coalescing random walks.
\newblock \emph{The Annals of Probability} 41(5): 3420--3461.

\bibitem[{Pasolini, Dardari, and Kieffer(2020)}]{pasolini2020exploiting}
Pasolini, G.; Dardari, D.; and Kieffer, M. 2020.
\newblock Exploiting the Agent's Memory in Asymptotic and Finite-Time Consensus
  Over Multi-Agent Networks.
\newblock \emph{IEEE Transactions on Signal and Information Processing over
  Networks} 6: 479--490.

\bibitem[{Tsitsiklis, Bertsekas, and Athans(1986)}]{tsitsiklis1986distributed}
Tsitsiklis, J.; Bertsekas, D.; and Athans, M. 1986.
\newblock Distributed asynchronous deterministic and stochastic gradient
  optimization algorithms.
\newblock \emph{IEEE transactions on automatic control} 31(9): 803--812.

\bibitem[{Tsitsiklis(1984)}]{tsitsiklis1984problems}
Tsitsiklis, J.~N. 1984.
\newblock \emph{Problems in Decentralized Decision Making and Computation}.
\newblock Ph.D. thesis, Massachusetts Institute of Technology.

\bibitem[{Yan, Jouandeau, and Cherif(2013)}]{yan2013survey}
Yan, Z.; Jouandeau, N.; and Cherif, A.~A. 2013.
\newblock A survey and analysis of multi-robot coordination.
\newblock \emph{International Journal of Advanced Robotic Systems} 10(12): 399.

\bibitem[{Zhong et~al.(2016)Zhong, Xu, Chen, Zhong, Qiu, Shi, and
  Wang}]{zhong2016generalized}
Zhong, L.-X.; Xu, W.-J.; Chen, R.-D.; Zhong, C.-Y.; Qiu, T.; Shi, Y.-D.; and
  Wang, L.-L. 2016.
\newblock A generalized voter model with time-decaying memory on a multilayer
  network.
\newblock \emph{Physica A: Statistical Mechanics and its Applications} 458:
  95--105.

\end{thebibliography}

\end{document}